%% file: recursiveBackwardsQlearning.tex
\documentclass{article}

\usepackage{arxiv}

\usepackage[utf8]{inputenc} 
\usepackage[T1]{fontenc}    
\usepackage{hyperref}       
\usepackage{url}            
\usepackage{booktabs}       
\usepackage{amsfonts}       
\usepackage{nicefrac}       
\usepackage{microtype}      
\usepackage{lipsum}		
\usepackage{graphicx}
\usepackage[backend=biber,
  isbn=false,                     
  autocite=inline,                
  style=ieee,                     
  hyperref=true,                  
  giveninits=false,               
  dashed=false                              
]{biblatex}
\usepackage{doi}
\usepackage{algorithmicx}
\usepackage{algpseudocode}
\usepackage{amsmath}
\usepackage{array}

\usepackage{xcolor}               
\usepackage{amsfonts}             
\usepackage{amssymb}              
\usepackage{acronym}
\usepackage{float}
\usepackage{algorithm}
\usepackage{pgfplots}
\usepackage{pgfplotstable}
\pgfplotsset{compat=1.7}
\usepackage{tikz}
\usetikzlibrary{positioning}
\usetikzlibrary{arrows}
\usepgfplotslibrary{fillbetween}
\title{Recursive Backwards Q-Learning in Deterministic Environments}

\author{\hspace{1mm}Jan Diekhoff \\
	Mannheim University\\
	of Applied Sciences\\
	Paul-Wittsack-Str. 10 \\
    68163 Mannheim \\
    Germany \\
	Email: jan.diekhoff@web.de
	\And
	\href{https://orcid.org/0000-0002-5102-3638}{\includegraphics[scale=0.06]{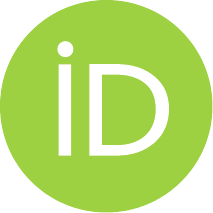}\hspace{1mm}Jörn Fischer} \\
	Mannheim University\\
	of Applied Sciences\\
	Paul-Wittsack-Str. 10 \\
    68163 Mannheim \\
    Germany \\
	Email: j.fischer@hs-mannheim.de
}

\date{}

\renewcommand{\headeright}{}
\renewcommand{\undertitle}{}
\renewcommand{\shorttitle}{Recursive Backwards Q-Learning in Deterministic Environments}

\hypersetup{
pdftitle={A template for the arxiv style},
pdfsubject={q-bio.NC, q-bio.QM},
pdfauthor={David S.~Hippocampus, Elias D.~Striatum},
pdfkeywords={First keyword, Second keyword, More},
}

\begin{document}
\maketitle

\begin{abstract}
	Reinforcement learning is a popular method of finding optimal solutions to complex problems. Algorithms like Q-learning excel at learning to solve stochastic problems without a model of their environment. However, they take longer to solve deterministic problems than is necessary. Q-learning can be improved to better solve deterministic problems by introducing such a model-based approach. This paper introduces the recursive backwards Q-learning (RBQL) agent, which explores and builds a model of the environment. After reaching a terminal state, it recursively propagates its value backwards through this model. This lets each state be evaluated to its optimal value without a lengthy learning process. In the example of finding the shortest path through a maze, this agent greatly outperforms a regular Q-learning agent.
\end{abstract}


\section{Introduction}
Machine learning and reinforcement learning are increasingly popular and important fields in the modern age. There are problems that reinforcement learning agents can learn to solve more efficiently and consistently than any human when given enough time to practice. However, modern approaches like Q-learning run into issues when facing certain types of problems. Their approach to solving problems in combination with not using a model of the environment causes them to take longer than is necessary to learn to solve problems that are deterministic in nature. By working without model of the environment, information that is available and help the learning process is ignored.

This paper introduces an adapted Q-learning agent called the \emph{recursive backwards Q-Learning (RBQL) agent}. It solves these types of problems by building a model of its environment as it explores and recursively applying the Q-value update rule to find an optimal policy much quicker than a regular Q-learning agent.
This agent is shown to work with the example of finding the fastest path through a maze. Its results are compared to the results of a regular Q-learning agent.

\section{Reinforcement Learning}
Reinforcement learning is one of the main fields of machine learning. It is commonly used for optimizing solutions to problems. At its most fundamental level, a reinforcement learning method is an implementation of an agent for solving a Markov decision process \cite{bellmanmarkov} by interacting with an environment. Markov decision processes describe problems as a set of states $S$, a set of actions $A$ and a set of rewards $R$. For every time step $t$, the agent chooses an action $a \in A$ and receives a new state $s \in S$ and a reward $r \in R$ for the action \cite{watkins}. Rewards may be positive or negative, depending on the outcome of the action, to encourage or discourage taking that action in the future \cite{suttonbarto}. The process of the agent interacting with the environment is called an episode which ends when a terminal state is reached which resets the environment and agent to their original configuration for the start of a new episode \cite{suttonbarto}. For the purposes of this paper, only finite Markov decision processes are considered, meaning the environment has at least one terminal state.

\begin{figure}[h]
\label{mdp}
    \centering
    \begin{tikzpicture}[scale=0.6]

      \node[rectangle,draw,scale=2,thick] (agent) {Agent};
      \node[rectangle,draw,below=8.75ex of agent, scale=2,thick] (environment) {Environment};
      
      \draw[dotted,scale=2,thick] -- (-2,-1.7) -- (-2,-2.4);
      
      \draw [-{latex},scale=2,thick] 
        (agent.east) -- +(2,0) |-  (environment);
      \node[scale=1.25,text width=1cm] at (7.5,-2) {action $A_t$};
    
      \draw [-{latex},scale=2,thick] 
        (environment.west)+(0,0.225) -- node[above] {$R_{t+1}$} (-2,-1.825);
        
      \draw [-{latex},scale=2,thick] 
        (environment.west)+(0,-0.225) -- node[below] {$S_{t+1}$} (-2,-2.275);
    
      \draw [-{latex},scale=2,thick] 
        (-2,-1.825) |- +(-.5,0) |- +(0,1.5725) -- (-.9,-0.25);
      \node[scale=1.25, text width=1cm] at (-3.8,-2) {reward $R_t$};    
        
      \draw [-{latex},scale=2,thick] 
        (-2,-2.275) |- +(-1,0) |- +(0,2.4725) -- (-.9,0.2);
      \node[scale=1.25, text width=1cm] at (-6.5,-2) {state $S_t$};
    \end{tikzpicture}
    \caption{Basic agent-environment relationship in a Markov decision process. The agent chooses an action $A_t$ and the environment returns a new state $S_{t+1}$ and a reward $R_{t+1}$. The dotted line represents the transition from step $t$ to step $t+1$ \cite{suttonbarto}.}
\end{figure}
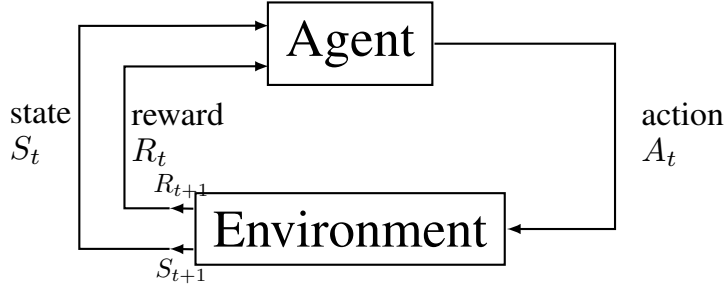

Reinforcement learning agents learn an optimal strategy for a given Markov decision process by estimating the value of either being in a state or taking a certain action in a certain state. They do this through a value function or action-value function respectively. The aim of the agent is to maximize the reward they receive in an episode \cite{suttonbarto}. To achieve this, value estimations do not only consider the immediate action the agent takes but also consider all future states and actions that may occur when taking the original action. Agents follow so-called policies according to which they choose which actions to take. Through gaining knowledge, they continuously adapt this policy in order to eventually reach an optimal policy - a policy which chooses the optimal action at every step. To explore, agents have to balance between exploration and exploitation \cite{suttonbarto}. Exploration is the act of following suboptimal actions to attempt to find an even better policy. On the other hand, exploitation is following the actions that will yield the currently highest estimated value. An agent that only exploits acts \emph{greedily}. To ensure continual exploration so that all actions get updated given enough time, agents can choose policies that are mostly greedy but choose to explore sometimes \cite{watkins}. To this end, an approach like $\epsilon$-greedy may be used. Here, $\epsilon$ is the probability of choosing a random action and $1-\epsilon$ is the probability of acting greedily.

A widely used modern approach to RL is temporal difference learning \cite{suttontd}, more specifically Q-learning \cite{watkins}. Q-learning works with the Q-learning update formula to update its policies:

\begin{equation}
\label{eq:qlearn}
\begin{aligned}
    Q(S_t,A_t) \leftarrow \ & Q(S_t,A_t) + \alpha \ \cdot \\ 
    & [R_{t+1} + \gamma\ \cdot \ \underset{a}{\text{max}}\ Q(S_{t+1},a) - Q(S_t,A_t)]
\end{aligned}
\end{equation}

$Q(S_t, A_t)$ is the estimated value for any given state-action pair. The equation shows how it is updated after taking action $A_t$ from state $S_t$. $R_{t+1}$ represents the reward gained, $\underset{a}{\text{max}}\ Q(S_{t+1},a)$ is the value estimation of the best action $a \in A_{t+1}$ that can be taken from $S_{t+1}$ according to the current policy, the state resulting from action $A_t$. $\alpha$ is a step-size parameter, also known as the \emph{learning rate}. Its value lies between $0$ and $1$ and it determines how importantly the agent values new information against the current estimate it already has. A value of $0$ completely ignores new information while a value of $1$ completely overrides the preexisting value estimate. $\gamma$ is the discount factor, weighing future rewards less than immediate ones. It also lies between $0$ and $1$, where $1$ weighs the best future action equally to the current one and $0$ does not consider it at all.

\section{Recursive Backwards Q-Learning}

\subsection{Idea}
\begin{figure}[h]
    \centering
    \begin{tabular}{c c c}
         & $Q$ & greedy policy \\
         & & w.r.t. $Q$ \\ \\
         ep. 0 & \input{diagramme/QProblem/q0} &  \input{diagramme/QProblem/greedy0} \\
         ep. 1 & \input{diagramme/QProblem/q1} &  \input{diagramme/QProblem/greedy1} \\
         ep. 2 & \input{diagramme/QProblem/q2} &  \input{diagramme/QProblem/greedy2} \\
         ep. 3 & \input{diagramme/QProblem/q3} &  \input{diagramme/QProblem/greedy3} \\
         ep. 4 & \input{diagramme/QProblem/q4} &  \input{diagramme/QProblem/greedy4} \\
         ep. 5 & \input{diagramme/QProblem/q5} &  \input{diagramme/QProblem/greedy5} \\
         ep. 6 & \input{diagramme/QProblem/q6} &  \input{diagramme/QProblem/greedy6}
    \end{tabular}
    \caption{Q-learning in a one-dimensional grid world. All Q-values are initialized as $-1$. Actions that lead to the terminal state reward 10. All other actions reward -1. The discount rate $\gamma$ is set to $0.9$. The learning rate $\alpha$ is set to $0.5$. The value of $\epsilon$ is irrelevant as the only action the agent takes is $\rightarrow$.}
    \label{fig:qproblem}
\end{figure}
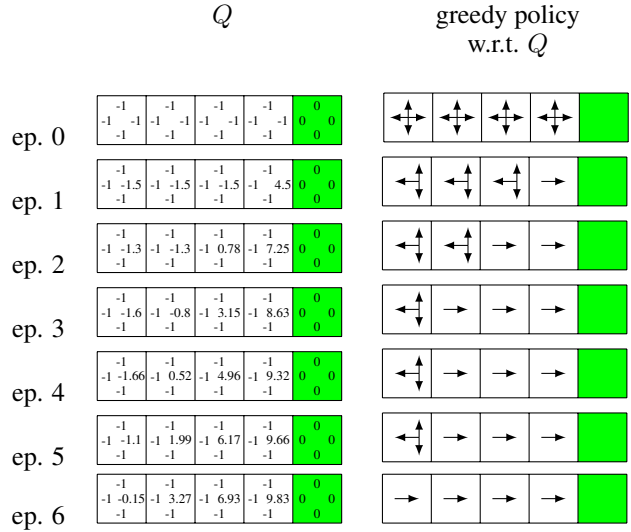

Q-learning agents are very widespread in modern reinforcement learning. Working free of a model allows them to be generally applicable to many problems. However, some Markov decision processes take longer to solve than is necessary because the agent ignores readily available information. This is noticeable in deterministic, episodic tasks where a positive reward is only given when reaching a terminal state. Before this state is reached for the first time, the agent appears to be moving entirely at random. Looking at figure \ref{fig:qproblem}, the issue becomes apparent. Even when following the optimal path at every step, it still takes multiple episodes for the reward of the terminal state to propagate back to the starting state. In fact, the optimal paths value estimation gets worse before it gets better. If every step has a reward of $-1$, values along the optimal path get worse if they do not lead to a state that has already been reached by the terminal state's positive reward as it travels backwards. 

In this paper, grid worlds \cite{suttonbarto} are used as an example Markov decision process for the agent to solve. Grid worlds are a two-dimensional grid in which every tile represents a state and the actions are limited to walking up, down, left or right. Grid worlds are  useful in that they are very simple to understand and to display, they have a limited set of actions and their set of states can be as small or large as is desired. Additionally, showing the value or optimal policy for each state is as easy as writing a number or drawing an arrow on the corresponding tile. Actions that would place the agent off of the grid simply return the state the agent is already in, but may still give a reward. Special tiles can also be defined, such as walls that act like the grid edge or pits that are terminal fail states because the agent cannot leave them once it has fallen in. Every grid world tile gives a reward of $-1$ to punish taking unnecessary actions in favor of taking the fastest path to the goal. 

Figure \ref{fig:qproblem} is a very simple grid world and it still takes six episodes to reach an optimal policy, even when taking the optimal action at every step. This problem will only grow worse and add noticeably more episodes of training for grid worlds that are not as trivial to solve, or even more complex tasks with more variables to consider. As stated, the issue is that the agent has no source of direction until it has randomly stumbled across the terminal state, its only source of positive rewards. The larger the state space, the longer it is blindly searching.

Reinforcement learning agents that work with a model of their environment are known as \emph{model-based} reinforcement learning agents. They can either work with a preexisting model or, more commonly, build their own. The way they construct their models is important as having perfect knowledge of an environment is neither feasible nor sensible. In the case of a grid world it is no problem, but imagining a more complex scenario like a self-driving car makes this fact apparent. When trying to drive from one city to another, knowing every centimeter of the road with every possible place other cars might be on the route is resource intensive and unnecessary. Instead, an agent should attempt to simplify its model as much as possible. Instead of every bit of road, long stretches going straight can be clumped together. Similar situations like a car in front slowing down can be treated the same wherever they occur.

The purpose of this paper is to introduce and evaluate a new type of model-based agent called the RBQL agent. The RBQL agent solves deterministic, episodic tasks that positively reward only the terminal state more efficiently than a regular Q-learning agent. It functions by building a model of its environment through exploration. When it reaches a terminal state, it recursively travels backwards through all previously explored states, applying a modified Q-learning update rule, the RBQL update rule. By setting the learning rate $\alpha$ to 1, equation (\ref{eq:qlearn}) can be simplified as such:

\begin{equation}
\label{eq:backwardsQ}
\begin{aligned}
    Q(S_t,A_t) \leftarrow & \ Q(S_t,A_t) + 1 \cdot [R_{t+1} + \gamma\ \underset{a}{\text{max}}\ Q(S_{t+1},a) \\
     & \ - Q(S_t,A_t)] \\
     = & \ Q(S_t,A_t) + R_{t+1} + \gamma\ \underset{a}{\text{max}}\ Q(S_{t+1},a) \\
     & \ - Q(S_t,A_t) \\
     = & \ R_{t+1} + \gamma\ \underset{a}{\text{max}}\ Q(S_{t+1},a)
\end{aligned}
\end{equation}
As can be seen in formula (\ref{eq:backwardsQ}), the Q-value now exclusively depends on the reward and the discounted reward of the best neighbor. Because the algorithm applies this formula starting with what is guaranteed to be the highest value of the environment and working its way away from it, the best possible neighbor for any given state is always the previously evaluated state. 

Evaluating all states at the end of the episode is reminiscent of dynamic programming \cite{bellmandp} or Monte Carlo methods \cite{suttonbarto} and is a point of critique for those approaches. However, as will be shown in chapter \ref{chap:results}, this evaluation method is so effective in RBQL that evaluating all known states in one go is still cost effective.
RBQL also differs in comparison to dynamic programming and Monte Carlo in a few major ways. In contrast with dynamic programming, it does not start out with a perfect model but has to build its own. It also propagates its reward throughout all states much more quickly and it uses an action-value function, not a state-value function. In contrast with Monte Carlo, it does not use exploring starts to guarantee exploration. It also does not only update the values that were seen in an episode. Instead, to facilitate exploration, it always prioritizes visiting unexplored actions, only following the greedy path when there are none. Because this mode of exploration still results in unexplored actions, the $\epsilon$-greedy approach is adapted for RBQL. Instead of exploring steps, the agent has exploration episodes. $\epsilon$ serves the same purpose as before, marking the probability of taking an exploration episode while $1-\epsilon$ is the probability of taking an exploitation episode. In an exploration episode, the agent randomly chooses an unexplored action anywhere in its model, navigates the model to put itself in a position to take that action and then continues to explore until it finds a known path again or the episode ends.

In this paper, finding an optimal path through a randomly generated grid world maze is used as an example task for RBQL to solve. It is also used to compare the performance of RBQL to Q-learning.

\subsection{Implementation}
\label{chap:implementation}
To implement RBQL\footnote{\ The source code can be downloaded at \url{https://github.com/JanDiekhoff/BackwardsLearner}}, the Godot game engine v. 3.5\footnote{\ Godot v. 3.5 can be downloaded at \url{https://godotengine.org/download/archive/3.5-stable/}} was used. Godot is a free, open source engine used mainly for video game development. Its main language is GDScript, an internal language that is very similar in syntax to Python, though it also supports C, C++, C\# and VisualScript. Because Python is very popular for machine learning development, the implementation is written in GDScript so that it is easily readable for interested parties. Godot uses a hierarchical structure of objects called \emph{nodes}. In the implementation, there are two main nodes: the agent and the environment. 

\subsubsection{Environment}
\label{chap:environment}
The environment is of the type \texttt{TileMap}\footnote{\url{https://docs.godotengine.org/en/3.5/classes/class_tilemap.html}} -- a class designed for creating maps in grid-based environments like grid worlds. Before starting the first episode, the environment generates a maze given a width $w$ and a height $h$ using a recursive backtracking algorithm \cite{rbm}. The starting point for the agent is always $(0,0)$ and the goal it attempts to reach -- the only terminal state -- is $(w-1,h-1)$. To ensure that the agent has the ability to improve even after finding the goal in the first episode, a maze with multiple paths is needed. Because a maze generated with recursive backtracking only has one path to the terminal state, a number of alternate paths are generated by taking $w \cdot h / 4$ random positions and a direction for each position. If the position has a wall in that direction, it is removed. If not, nothing happens.

The environment has a function \texttt{step(state,action)} that serves as the only way for the agent to interact with it. The possible moves are \texttt{UP}, \texttt{DOWN}, \texttt{LEFT} and \texttt{RIGHT}. The state is described as a coordinate of the current position. In Godot, the class \texttt{Vector2(x,y)
}\footnote{\url{https://docs.godotengine.org/en/3.5/classes/class_vector2.html}} is used for this purpose. \texttt{step()} checks if taking the given action from the given state results in hitting a wall or not. If not, the agent moves to a new position. There are three different rewards: $-1$ for any normal tile, $-5$ for hitting a wall and $10$ for reaching the terminal state. $-1$ is awarded at every step to discourage agents from taking unnecessary steps. Walls give $-5$ to quickly teach the agent to ignore them. After taking an action, the new state and reward are returned to the agent, as well as a notification if the episode has ended or not and if the agent has hit a wall or not.

The \texttt{TileMap} has a tile for each combination of having or not having a wall in each of the four directions, totaling $2^4$ or $16$ total possible tiles. Another option would be to just have a floor tile and a wall tile. However, that would make a maze with an equivalent wall layout much larger, leading to a larger state set and longer solving times. To determine if a wall is in a certain direction, the id of each tile from $0$ to $15$ acts as a four-bit flag. Each direction is assigned one of the bits ($\texttt{UP} = 0$, $\texttt{RIGHT} = 1$, $\texttt{DOWN} = 2$ and $\texttt{LEFT} = 3$). If the flag is set, there is a wall in the corresponding direction. The id for an L-shaped tile for example would be $2^2 + 2^3 = 12$ as \texttt{DOWN} and \texttt{LEFT} have walls. The process for determining if the agent can move in a given direction $d$ from a position $p$ is $(\neg id_p) \And (2^d)$, where $id_p$ is the id of the tile at $p$.

\subsubsection{RBQL Agent}
\label{chap:rbqlagent}
The RBQL agent is represented by a \texttt{Sprite}\footnote{\url{https://docs.godotengine.org/en/3.5/classes/class_sprite.html}} object -- a 2D image -- so it can be observed while solving a maze. During its runtime, the agent keeps track of a few key things:
\begin{itemize}
    \item A model of the environment (\texttt{explored\_map})
    \item A list of rewards for each state-action pair (\texttt{rewards})
    \item The last reward received (\texttt{reward})
    \item A list of steps taken per episode (\texttt{steps\_taken})
    \item The Q-table (\texttt{qtable})
    \item The current state (\texttt{current\_state})
    \item The previous state (\texttt{old\_state})
    \item The last taken action (\texttt{action})
\end{itemize}

The model of the environment starts out as an empty dictionary. Every time a new state is discovered, an entry for that state is made and initialized as an empty array. When an action is taken from this state, the resulting new state is entered into the previous state's array at the index of the taken action's designated number (\texttt{explored\_map[old\_state][action] = current\_state}). When hitting a wall, the ``new'' state is the same as the state from which the action was taken. Similarly, when an action is taken, the resulting reward is saved in the rewards list (\texttt{rewards[old\_state][action] = reward}). Because the agent uses state-action values, not state values, the tiles are treated like nodes in a directed graph. Going from tile A to tile B might result in a different reward than when going from B to A, so when the agent learns the reward of going from A to B, it does not also learn the reward of going from B to A.

Being a Q-learner makes it simpler to generalize the agent for other tasks, but it causes a lot of exploratory steps and exploratory episodes to only explore one position at a time. If an exploration episode chooses an unexplored state-action pair that results in hitting a wall, the exploration episode immediately ends with little information gained. To alleviate this problem, the agent takes exploratory ``look-ahead'' steps. After entering a tile, it takes a step in every direction but only saves the result if it hits a wall. This guarantees that exploratory episodes always take new paths and not just hit a wall and continue on the best known path.

The agent also keeps track of a list of the actions it has taken -- except for when hitting a wall -- for the case that it reaches a dead end, or rather a state with no unexplored neighbors. In this case, the agent would normally follow the optimal path until it finds a new unexplored path or reaches the terminal state. However, if the path the agent is on has not been explored before it has not yet been evaluated and there is no optimal path to follow. In this case, the agent backtracks by taking the opposite action of the most recent in the list, then removes it from the list, until an unexplored tile or an evaluated path to follow is found.

Finally, when the terminal state is reached, the Q-table is updated with the rewards saved in \texttt{rewards} according to the RBQL update rule.
\begin{equation*}
\begin{aligned}
    & \texttt{qtable[state][action] = } \\
    & \texttt{rewards[state][action] + discount\_rate $\cdot$ } \\
    & \texttt{qtable[explored\_map[state][action]].max()}
\end{aligned}
\end{equation*}

To do this, a copy of \texttt{explored\_map} is inverted to be able to traverse it in reverse. This is then done with a breadth-first search algorithm, starting at the terminal state, and the Q-value is calculated for each state. Breadth-first search is chosen over a depth-first search algorithm so that each state must only be visited once as the value is directly proportional to the distance from the terminal state. With breadth-first search, each state gets the highest possible value on its first visit because it is visited from its highest possible valued neighbor.

When all known states have been evaluated, a new episode begins. After the first episode, episodes are chosen to be either exploratory or exploitative, similar to how an $\epsilon$-greedy policy may choose exploratory actions. In an exploitative episode, the agent simply follows the best path it knows, choosing at random if two states are equally good, but still always exploring unknown states directly adjacent to the path above all else. In an exploratory episode, a random state with an unexplored neighbor is chosen. The agent navigates to this state with the help of the A* search algorithm \cite{astar} and follow the unexplored path from there until it finds a known state again. This exploratory excursion may only find one new state or it may find a vastly superior path to what was known before. $\epsilon$ is decreased after every episode as follows:
\begin{equation*}
\begin{aligned}
\epsilon =\ & \texttt{min\_epsilon + (max\_epsilon - min\_epsilon)} \\
&  \texttt{$\cdot$} \ \   e^{(-\texttt{decay\_rate $\cdot$ current\_episode})}
\end{aligned}
\end{equation*}
where \texttt{min\_epsilon}, \texttt{max\_epsilon} and \texttt{decay\_rate} can be any value within a range of $[0,1]$ and \texttt{current\_episode} is the number of the current episode starting with $0$. Once every state is explored, the agent is guaranteed to have found the optimal path, or paths, through the maze.
In its entirety, the algorithm can be expressed like this:

\begin{algorithm}[H]
\label{alg:backwardsQ}
\caption{Backwards Q-Learning Algorithm}
\begin{algorithmic}
    \State Set exploration\_episode to false
    \While{true}
        \If{exploration\_episode}
            \State Find unexplored path
            \State Travel to unexplored path
        \EndIf
        \While{episode is not over}
            \If{current position has an unexplored neighbor}
                \State Visit unexplored neighbor
                \State Update model
                \State Save reward
                \If{no wall hit}
                    \State Save action in action queue
                \EndIf
            \ElsIf{there is an optimal path to follow}
                \State Visit best neighbor
            \EndIf
            \While{current pos. has no unexplored neighbor}
                \State Backtrack
            \EndWhile
        \EndWhile
        
        \State Create state queue with breadth-first search
        \For{state in queue}
            \State Apply RBQL formula
        \EndFor
        \State Set exploration\_episode to random() $<= \epsilon$
        \State Apply decay to $\epsilon$
    \EndWhile
\end{algorithmic}
\end{algorithm}

\subsubsection{Q-learning agent}
\label{chap:qagent}
A standard Q-learning agent has been implemented in Godot as well to compare the performance of the RBQL agent to. This agent is comparatively simple:

\begin{algorithm}
\label{alg:Q}
\caption{Q-Learning Algorithm}
\begin{algorithmic}
    \While{true}
        \If{random() $<= \epsilon$}
            \State Choose random action
        \Else
            \State Choose greedy action
        \EndIf
        \State Take action
        \State Receive new state and reward
        \State Update Q-table for old state and action
        \If{terminal state reached}
            \State Start new episode
        \EndIf
        \State Apply decay to $\epsilon$
    \EndWhile
\end{algorithmic}
\end{algorithm}

\section{Tests and Results}
\label{chap:results}
To compare the performance of the two agents, three sets of tests have been done for different maze sizes: $5\times5$, $10\times10$ and $15\times15$. All variables have been set to common values. The decay rate is set somewhat high to account for the relatively low episode amount: 
\begin{itemize}
    \item $\gamma = 0.9$
    \item $\alpha = 0.1$ (RBQL has $\alpha = 1$ as explained in equation (\ref{eq:backwardsQ}))
    \item $\texttt{min\_epsilon} = 0.01$
    \item $\texttt{max\_epsilon} = 1$
    \item $\texttt{decay\_rate} = -0.01$
\end{itemize}
For every maze size, each agent is given the same set of 50 randomly generated mazes. Each agent is given 25 episodes per maze to train. These values are chosen to offer a reasonably large sample size without requiring an enormous amount of time to compute. Agents are compared by the number of steps taken per episode, with less steps taken being a more desirable outcome. The step counter is increased every time \texttt{step()} is called, including the look-ahead steps of the RBQL. For a sense of perspective, the best possible solution to any square maze of size $s^2$ is $2s-2$. Assuming a maze with no walls, the shortest distance between two points $A$ and $B$ can be expressed as their Manhattan distance $|A_X - B_X| + |A_Y - B_Y|$ \cite{taxicab}. In the corners of a square, it holds that $A_X = A_Y$ and $B_X = B_Y$, so the distance can be simplified as $2\cdot|A-B|$. Setting $A = 0$ and $B = s-1$, this further simplifies to $2s-2$. This means that while the amount of states (and thereby state-action pairs) increases quadratically, the best possible solution only increases linearly. This in turn means that the amount of states that are not on the optimal path that the agent has to evaluate will often increase drastically with the size of the maze.

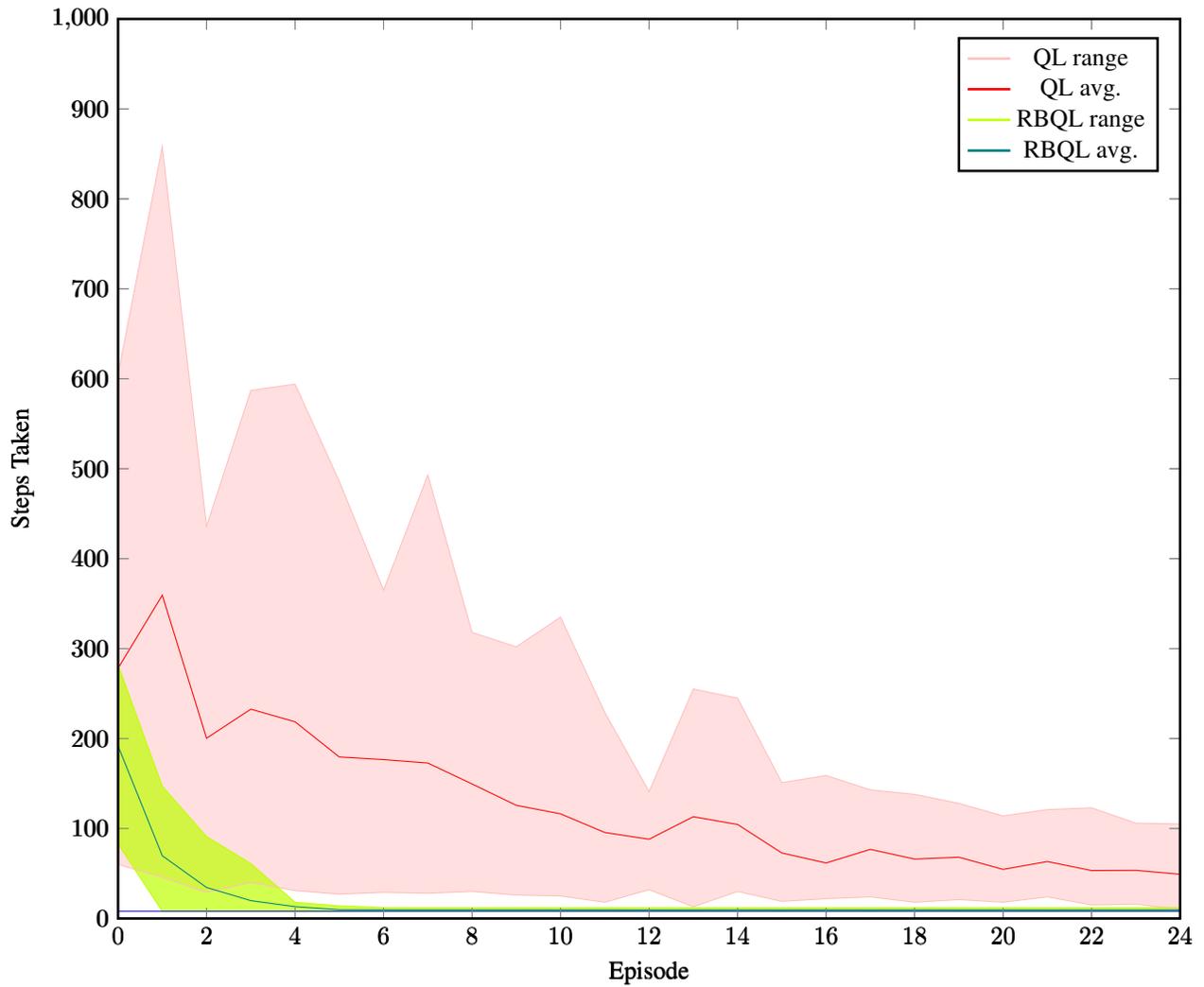
\begin{figure}
    \centering
    \begin{tikzpicture}
        \begin{axis}[ 
            width=\linewidth,
            line width=1,
            tick label style={font=\normalsize},
            legend style={nodes={scale=1, transform shape}},
            xlabel={Episode},
            ylabel={Steps Taken},
            y tick label style={
            /pgf/number format/.cd,
            precision=1
            },
            ymin=0,ymax=1000,
            xmin=0,xmax=24
        ]
        \addplot[pink] coordinates{(0,0)};
        \addlegendentry{QL range};
        \addplot[red] coordinates{(0,0)};
        \addlegendentry{QL avg.};
        \addplot[lime] coordinates{(0,0)};
        \addlegendentry{RBQL range};
        \addplot[teal] coordinates{(0,0)};
        \addlegendentry{RBQL avg.};
            
        \addplot[pink, name path=f] coordinates {
        (0,605)(1,858)(2,436)(3,587)(4,594)(5,486)(6,365)(7,493)(8,318)(9,302)(10,335)(11,229)(12,141)(13,255)(14,245)(15,151)(16,159)(17,143)(18,138)(19,128)(20,114)(21,121)(22,123)(23,106)(24,105)
        };
        \addplot[pink, name path=g] coordinates {
        (0,60)(1,46)(2,29)(3,40)(4,31)(5,27)(6,29)(7,28)(8,30)(9,26)(10,25)(11,18)(12,32)(13,13)(14,30)(15,19)(16,22)(17,24)(18,18)(19,21)(20,18)(21,24)(22,15)(23,16)(24,10)
        };
        \addplot[pink, opacity=0.5] fill between[of=f and g];

        \addplot[red] coordinates {(0,278.06)(1,359.44)(2,200.36)(3,232.7)(4,218.64)(5,179.54)(6,176.62)(7,172.82)(8,149.54)(9,125.8)(10,116.3)(11,95.58)(12,88.1)(13,113.14)(14,104.5)(15,72.82)(16,61.66)(17,76.84)(18,66.02)(19,68.08)(20,54.6)(21,63.28)(22,53.3)(23,53.5)(24,49.14)};

        \addplot[lime, name path=h] coordinates {
        (0,282)(1,147)(2,91)(3,61)(4,18)(5,14)(6,12)(7,12)(8,12)(9,12)(10,12)(11,12)(12,12)(13,12)(14,12)(15,12)(16,12)(17,12)(18,12)(19,12)(20,12)(21,12)(22,12)(23,12)(24,12)
        };
        \addplot[lime, name path=i] coordinates {
        (0,83)(1,8)(2,8)(3,8)(4,8)(5,8)(6,8)(7,8)(8,8)(9,8)(10,8)(11,8)(12,8)(13,8)(14,8)(15,8)(16,8)(17,8)(18,8)(19,8)(20,8)(21,8)(22,8)(23,8)(24,8)
        };
        \addplot[lime, opacity=0.7] fill between[of=h and i];
        
        \addplot[blue] coordinates {(0,8)(24,8)};
        
        \addplot[teal] coordinates {(0,191.84)(1,69.74)(2,34.48)(3,19.84)(4,13.1)(5,9.74)(6,9.64)(7,9.64)(8,9.64)(9,9.64)(10,9.62)(11,9.64)(12,9.56)(13,9.62)(14,9.64)(15,9.62)(16,9.64)(17,9.62)(18,9.64)(19,9.62)(20,9.62)(21,9.62)(22,9.64)(23,9.64)(24,9.62)};
        \end{axis}
        \begin{axis}[ 
            width=\linewidth,
            line width=1,
            tick label style={font=\normalsize},
            legend style={nodes={scale=1, transform shape}},
            xlabel={Episode},
            ylabel={Steps Taken},
            y tick label style={
            /pgf/number format/.cd,
            precision=1
            },
            ymin=0,ymax=1000,
            xmin=0,xmax=24
        ]
        \end{axis}
    \end{tikzpicture}
    \caption{Number of steps taken to find the goal in a randomly generated grid world maze of size $5\times5$. The blue line is the minimum step threshold for any maze of this size. The light red area shows the range of Q-learning agent's highest and lowest step count, excluding the highest and lowest two. The red line shows the average performance. Similarly, the light green area shows the range of the RBQL agent's highest and lowest step count, excluding the highest and lowest two, and the green line shows the average performance.}
    \label{fig:fivebyfive}
\end{figure}

\begin{figure}
    \centering
    \begin{tikzpicture}
        \begin{axis}[ 
            width=\linewidth,
            line width=1,
            tick label style={font=\normalsize},
            legend style={nodes={scale=1, transform shape}},
            xlabel={Episode},
            ylabel={Steps Taken},
            y tick label style={
            /pgf/number format/.cd,
            precision=1
            },
            ymin=0,ymax=8000,
            xmin=0,xmax=24
        ]
        \addplot[pink] coordinates{(0,0)};
        \addlegendentry{QL range};
        \addplot[red] coordinates{(0,0)};
        \addlegendentry{QL avg.};
        \addplot[lime] coordinates{(0,0)};
        \addlegendentry{RBQL range};
        \addplot[teal] coordinates{(0,0)};
        \addlegendentry{RBQL avg.};
        
        \addplot[pink, name path=f] coordinates {
        (0,7585)(1,5177)(2,5079)(3,4013)(4,3817)(5,3501)(6,2917)(7,2551)(8,2153)(9,1984)(10,1852)(11,1699)(12,1633)(13,1764)(14,771)(15,1237)(16,1038)(17,746)(18,912)(19,894)(20,555)(21,593)(22,681)(23,739)(24,642)
        };
        \addplot[pink, name path=g] coordinates {
        (0,487)(1,442)(2,267)(3,309)(4,227)(5,346)(6,215)(7,236)(8,171)(9,151)(10,156)(11,155)(12,147)(13,133)(14,89)(15,140)(16,117)(17,117)(18,80)(19,85)(20,92)(21,81)(22,125)(23,64)(24,73)
        };
        \addplot[pink, opacity=0.5] fill between[of=f and g];
        
        \addplot[red] coordinates {(0,3308.46)(1,2384.82)(2,2507.54)(3,1818.14)(4,1730.12)(5,1296.68)(6,1126.9)(7,1066.9)(8,992.1)(9,843.86)(10,744.66)(11,756.72)(12,698.34)(13,656.22)(14,379.8)(15,458.78)(16,435.6)(17,363.34)(18,371.42)(19,371.78)(20,346.16)(21,286.0)(22,317.0)(23,293.12)(24,281.44)};

        \addplot[lime, name path=h] coordinates {
        (0,1252)(1,710)(2,517)(3,356)(4,173)(5,82)(6,32)(7,32)(8,32)(9,32)(10,32)(11,32)(12,32)(13,32)(14,32)(15,32)(16,32)(17,32)(18,32)(19,32)(20,32)(21,32)(22,32)(23,32)(24,32)
        };
        \addplot[lime, name path=i] coordinates {
        (0,376)(1,40)(2,20)(3,20)(4,18)(5,18)(6,18)(7,18)(8,18)(9,18)(10,18)(11,18)(12,18)(13,18)(14,18)(15,18)(16,18)(17,18)(18,18)(19,18)(20,18)(21,18)(22,18)(23,18)(24,18)
        };
        \addplot[lime, opacity=0.7] fill between[of=h and i];
        
        \addplot[blue] coordinates {(0,18)(24,18)};
        
        \addplot[teal] coordinates {(0,843.52)(1,218.58)(2,150.64)(3,117.04)(4,61.82)(5,54.08)(6,25.1)(7,35.5)(8,23.76)(9,23.68)(10,23.68)(11,23.68)(12,23.68)(13,23.68)(14,23.68)(15,23.68)(16,23.68)(17,23.68)(18,23.68)(19,23.68)(20,23.68)(21,23.68)(22,23.68)(23,23.68)(24,23.68)};

        \end{axis}
        \begin{axis}[ 
            width=\linewidth,
            line width=1,
            tick label style={font=\normalsize},
            legend style={nodes={scale=1, transform shape}},
            xlabel={Episode},
            ylabel={Steps Taken},
            y tick label style={
            /pgf/number format/.cd,
            precision=1
            },
            ymin=0,ymax=8000,
            xmin=0,xmax=24
        ]
        \end{axis}
    \end{tikzpicture}
    \caption{Number of steps taken to find the goal in a randomly generated grid world maze of size $10\times10$. The light red area shows the range of Q-learning agent's highest and lowest step count, excluding the highest and lowest two. The red shows the average performance. Similarly, the light green area shows the range of the RBQL agent's highest and lowest step count, excluding the highest and lowest two, and the green line shows the average performance.}
    \label{fig:tenbyten}
\end{figure}

\begin{figure}
    \centering
    \begin{tikzpicture}
        \begin{axis}[ 
            width=\linewidth,
            line width=1,
            tick label style={font=\normalsize},
            legend style={nodes={scale=1, transform shape}},
            xlabel={Episode},
            ylabel={Steps Taken},
            ymin=0,ymax=22000,
            xmin=0,xmax=24
        ]
        \addplot[pink] coordinates{(0,0)};
        \addlegendentry{QL range};
        \addplot[red] coordinates{(0,0)};
        \addlegendentry{QL avg.};
        \addplot[lime] coordinates{(0,0)};
        \addlegendentry{RBQL range};
        \addplot[teal] coordinates{(0,0)};
        \addlegendentry{RBQL avg.};
        
        \addplot[pink, name path=f] coordinates {
        (0,17116)(1,21147)(2,14494)(3,10330)(4,12760)(5,10774)(6,8489)(7,7787)(8,5282)(9,4168)(10,3532)(11,3865)(12,3129)(13,3569)(14,4749)(15,2467)(16,2033)(17,2505)(18,2053)(19,2135)(20,1616)(21,2123)(22,1452)(23,1952)(24,1680)
        };
        \addplot[pink, name path=g] coordinates {
        (0,667)(1,507)(2,1014)(3,928)(4,940)(5,822)(6,675)(7,692)(8,323)(9,425)(10,377)(11,553)(12,398)(13,458)(14,319)(15,306)(16,288)(17,356)(18,237)(19,244)(20,264)(21,263)(22,268)(23,295)(24,189)
        };
        \addplot[pink, opacity=0.5] fill between[of=f and g];
        
        \addplot[red] coordinates {(0,7180.98)(1,7808.02)(2,6013.14)(3,4881.78)(4,5097.96)(5,4171.18)(6,3679.2)(7,2948.72)(8,2266.84)(9,1730.02)(10,1735.58)(11,1799.7)(12,1386.92)(13,1457.4)(14,1652.16)(15,1159.82)(16,1185.06)(17,1116.94)(18,1094.98)(19,1013.7)(20,843.16)(21,963.6)(22,727.28)(23,852.44)(24,778.68)};

        \addplot[lime, name path=h] coordinates {
        (0,2855)(1,1640)(2,759)(3,659)(4,366)(5,302)(6,201)(7,124)(8,100)(9,141)(10,60)(11,56)(12,85)(13,46)(14,50)(15,46)(16,46)(17,46)(18,46)(19,46)(20,46)(21,46)(22,46)(23,46)(24,46)
        };
        \addplot[lime, name path=i] coordinates {
        (0,949)(1,62)(2,36)(3,32)(4,30)(5,30)(6,30)(7,30)(8,30)(9,30)(10,30)(11,30)(12,30)(13,30)(14,30)(15,30)(16,30)(17,30)(18,30)(19,30)(20,30)(21,30)(22,30)(23,30)(24,30)
        };
        \addplot[lime, opacity=0.7] fill between[of=h and i];
        
        \addplot[blue] coordinates {(0,28)(24,28)};

        \addplot[teal] coordinates {(0,1965.0)(1,506.18)(2,243.0)(3,175.32)(4,140.92)(5,108.8)(6,91.62)(7,65.58)(8,50.76)(9,53.74)(10,48.28)(11,40.3)(12,45.36)(13,35.96)(14,37.32)(15,35.96)(16,35.96)(17,35.96)(18,35.96)(19,35.96)(20,35.96)(21,35.96)(22,35.96)(23,35.96)(24,35.96)};
        \end{axis}
        
        \begin{axis}[ 
            width=\linewidth,
            line width=1,
            tick label style={font=\normalsize},
            legend style={nodes={scale=1, transform shape}},
            xlabel={Episode},
            ylabel={Steps Taken},
            ymin=0,ymax=22000,
            xmin=0,xmax=24
        ]
        \end{axis}
    \end{tikzpicture}
    \caption{Number of steps taken to find the goal in a randomly generated grid world maze of size $15\times15$. The light red area shows the range of Q-learning agent's highest and lowest step count, excluding the highest and lowest two. The red line shows the average performance. Similarly, the light green area shows the range of the RBQL agent's highest and lowest step count, excluding the highest and lowest two, and the green line shows the average performance.}
    \label{fig:fiftenbyfiften}
\end{figure}

Looking at the results, a few things can be observed. First of all, the average number of steps the RBQL agent takes is consistently lower than the Q-learning agent in all three maze sizes. It also has much less variation in step counts, which can be seen when looking at the areas of lighter hue. The light red areas are much more sporadic and spike further away from the average. The green areas stick much closer together. If the highest two step counts per episode were not removed, RBQL would also have a few small spikes. These spikes would represent exploratory episodes where a new path is explored, resulting in a higher step count. In cases where the line is flat for a long period of time, it can be assumed that the optimal solution is found. This can be seen in all three figures, where both the average and the min/max range become a straight line close to the minimum. Important to note is that every maze has a different optimal solution, hence why the average sits above the blue line which denotes the lowest possible step count in any maze of this size. It can also be observed that none of the lines ever go below this boundary, as is to be expected.

Second, even when removing the highest two step counts per episode, many of the Q-learning agent's step counts are so large that scaling the graphs to fit them makes the RBQL agent's data and the lower boundary difficult to see in the graphs for the larger mazes. The highest step count values that have not been cut are 858 steps in figure \ref{fig:fivebyfive}, 7,585 in figure \ref{fig:tenbyten} and 21,147 in figure \ref{fig:fiftenbyfiften}, while the highest in total are 3,716 steps in figure \ref{fig:fivebyfive}, 20,553 in figure \ref{fig:tenbyten} and 26,315 in figure \ref{fig:fiftenbyfiften}.

Third, it is interesting to see how the differences in average step counts evolve with the grid size. Table \ref{fig:diff} shows this difference in the first and last episode. The difference between the average step counts in the last episode especially is striking, as it is close to doubling from each size to the next. Further, looking at the improvement of each agent as seen in table \ref{fig:improvement}, one can see that the factor by which RBQL improves massively increases the bigger the maze becomes while the Q-learner only slightly improves its performance in comparison. Additionally, most of the improvement of RBQL is done in the first two episodes, while the Q-learner has a more gradual learning curve.

\begin{table}[H]
    \centering
    \caption{Difference in average step counts of the Q-learner and RBQL. The difference expresses how many times more steps the Q-learner took compared to RBQL.}
    \begin{tabular}{cccc}
        \\
        \ Grid size\  & \ Q-learner steps\  & \ RBQL steps\  & \ Difference\ \\\hline
        \multicolumn{4}{c}{\textbf{Episode 0}}\\\hline
        $5\times5$ & 278.06 & 191.84 & 1.45 \\\hline
        $10\times10$ & 3,308.46 & 843.52 & 3.92 \\\hline
        $15\times15$ & 7,180.98 & 1,965 & 3.65 \\\hline
        \multicolumn{4}{c}{\textbf{Episode 24}}\\\hline
        $5\times5$ & 49.14 & 9.62 & 5.11 \\\hline
        $10\times10$ & 281.44 & 23.68 & 11.89 \\\hline
        $15\times15$ & 778.68 & 35.96 & 21.65  \\
    \end{tabular}
    \label{fig:diff}
\end{table}

Lastly, the RBQL agent seems to find an optimal policy at around episode 4 for the $5\times5$, episode 6 for the $10\times10$ and episode 10 for the $15\times15$ grid. As the previous figures show, the Q-learning agent does not come close to similarly low step counts and therefore does not reach an optimal policy at all with the same amount of training.

\begin{table}[H]
    \centering
    \caption{Difference in average step counts of the Q-learner and RBQL. Improvement shows the factor by which the amount of steps is reduced from episode 0 to 24.}
    \begin{tabular}{cccc}
        \\
        \ Grid size\ & \ Steps in episode 0\ & \ Steps in episode 24\  & \ Improvement\ \\\hline
        \multicolumn{4}{c}{\textbf{Q-learning agent}}\\\hline
        $5\times5$ & 278.06  & 49.14 & 5.66 \\\hline
        $10\times10$ & 3,308.46 & 281.44 & 11.76 \\\hline
        $15\times15$ & 7,180.98 & 778.68 & 9.22 \\\hline
        \multicolumn{4}{c}{\textbf{RBQL agent}}\\\hline
        $5\times5$ & 191.84  & 9.62 & 19.94 \\\hline
        $10\times10$ & 843.52 & 23.68 & 35.62 \\\hline
        $15\times15$ & 1,965 & 35.96 & 90.76  \\
    \end{tabular}
    \label{fig:improvement}
\end{table}

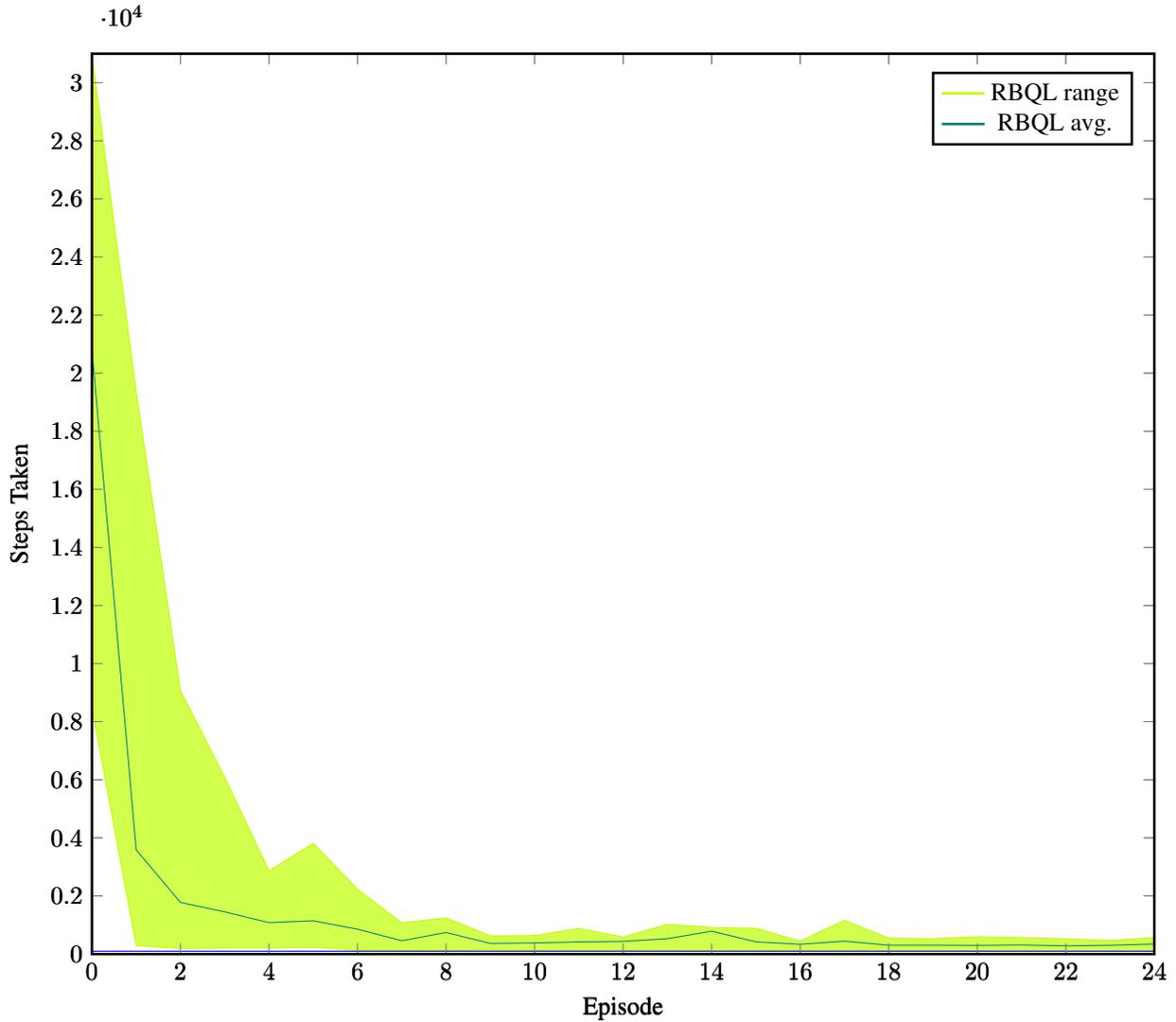
\begin{figure}
    \centering
    \begin{tikzpicture}
        \begin{axis}[ 
            width=\linewidth,
            line width=1,
            tick label style={font=\normalsize},
            legend style={nodes={scale=1, transform shape}},
            xlabel={Episode},
            ylabel={Steps Taken},
            ymin=0,ymax=31000,
            xmin=0,xmax=24
        ]
        
        \addplot[lime] coordinates{(0,0)};
        \addlegendentry{RBQL range};
        \addplot[teal] coordinates{(0,0)};
        \addlegendentry{RBQL avg.};
        
        \addplot[lime, name path=h] coordinates {
        (0,30947)(1,19358)(2,9072)(3,6081)(4,2864)(5,3810)(6,2232)(7,1075)(8,1246)(9,626)(10,635)(11,882)(12,584)(13,1023)(14,916)(15,884)(16,446)(17,1160)(18,546)(19,525)(20,596)(21,563)(22,524)(23,468)(24,565)
        };
        \addplot[lime, name path=i] coordinates {
        (0,8478)(1,295)(2,187)(3,205)(4,204)(5,228)(6,134)(7,130)(8,128)(9,140)(10,120)(11,126)(12,130)(13,134)(14,122)(15,122)(16,130)(17,122)(18,122)(19,122)(20,122)(21,122)(22,114)(23,116)(24,118)
        };
        \addplot[lime, opacity=0.7] fill between[of=h and i];
        
        \addplot[blue] coordinates {(0,98)(24,98)};
        
        \addplot[teal] coordinates {(0,20811.08)(1,3586.98)(2,1777.44)(3,1451.74)(4,1081.0)(5,1145.2)(6,855.68)(7,458.62)(8,742.4)(9,365.42)(10,381.38)(11,416.74)(12,435.14)(13,528.28)(14,787.32)(15,420.3)(16,336.34)(17,444.04)(18,303.8)(19,306.08)(20,296.3)(21,313.06)(22,281.12)(23,300.7)(24,344.9)};

        \end{axis}
        \begin{axis}[ 
            width=\linewidth,
            line width=1,
            tick label style={font=\normalsize},
            legend style={nodes={scale=1, transform shape}},
            xlabel={Episode},
            ylabel={Steps Taken},
            ymin=0,ymax=31000,
            xmin=0,xmax=24
        ]
        \end{axis}
    \end{tikzpicture}
    \caption{Number of steps taken to find the goal in a randomly generated grid world maze of size $50\times50$. The light green area shows the range of the RBQL agent's highest and lowest step count, excluding the highest and lowest two, and the green line shows the average performance.}
    \label{fig:fiftybyfifty}
\end{figure}

To further show RBQL's efficiency, it has also been tested under the same parameters in a grid of size $50\times50$. The results can be seen in figure \ref{fig:fiftybyfifty}. This test is done to demonstrate that even such a large maze can be explored by RBQL. As with the previous examples, by far the largest policy improvement still happens in the first episode. With mazes of such a large size, a lot more spikes in step counts are seen in later episodes because there are more states to explore. The difference in average step counts goes from 20,811.08 in episode 0 to 344.9 in episode 24, an improvement by a factor of 60.34. This is worse than the improvement in the $15\times15$ mazes, but still almost double that of the $10\times10$ mazes.

\section{Discussion}
\label{chap:discussion}
This chapter explores the practicality of using this algorithm to solve other Markov decision processes. It discusses which parts of the implementation are and are not specific to the problem of fastest path through a maze, which improvements can be made to make it more applicable for other problems and showcases further points for research in this field. The constraints given in this paper are that the agent will attempt to solve deterministic, episodic tasks with a single terminal state as its only source of positive rewards. This chapter also discusses which of these constraints can be dismissed.

There are a few parts of the implementation as presented in chapter \ref{chap:implementation} that are only applicable to this specific problem. This is not necessarily a bad thing, as the purpose of the RBQL agent is to utilize knowledge of its environment. As a result of this, the only parts that cannot be directly adapted for other problems are the way the agent builds its model. In the grid world maze, it can assume that every state has the same actions it can take and has a neighboring state in each direction (though it may sometimes be itself). Further, every action always has an opposite action, going up can always be undone by going down for example. These assumptions allow it to easily build a model of the grid world and influence its policy in how it further explores it. They allow the agent to take steps in each direction to check for walls and they allow the agent to backtrack when it is stuck in a dead end. These assumptions cannot be guaranteed for other Markov decision processes or even for grid worlds with more complex behavior like a wind tunnel that if walked through also pushes the agent one tile in the direction the wind is traveling. The way in which the agent builds a model has to either be designed for each environment individually or it has to be abstracted so that it is more broadly applicable. Finding such an approach to model building is one area of improvement for RBQL. Importantly though, none of these assumptions are required for the agent to function. Backtracking, opposite steps and the same actions for every state simply make the implementation easier and more efficient. As long as no path of a directed graph would cause the agent to be stuck with no way to reach a terminal state, it can be explored and evaluated.

Another improvement to the way the implementation builds its model is to simplify it as far as possible. As the amount of states directly influences how long a problem takes to solve, RBQL will become more efficient the more it can remove unnecessary states. Currently, every position has its own state. If the agent could detect ``hallways'' -- tiles with parallel walls -- they could be removed without problem in favor of directly connecting the two tiles at either side of the hallway -- only the negative rewards for the length of the hallway would have to be implemented into the model. Further, if there is a non-forking path that leads into a dead end, the entire path could be treated as a wall and ignored entirely. This would leave only the starting state, terminal state, turns and forking paths to evaluate. Both of these additions leave the key part of the algorithm, traversing the model backwards and applying the RBQL update formula, untouched.

RBQL can be easily adapted to include multiple terminal states with the same or different rewards and this is already supported by the implementation. There are two possible ways to do this. First is to create an imaginary state that all terminal states lead into from which the backtracking always starts. Second is to remember all terminal states and backtrack from each of them. The first option is much more efficient as each state still only gets evaluated once while the second version avoids having to tamper with the model. 

Finally, RBQL could be adapted to work in non-deterministic environments. To reiterate, deterministic means that a state-action pair always yields the same state-reward pair. If the agent could, while building its model, also estimate the transition probabilities of a state-action pair to a new state, RBQL could still be used to evaluate the states. The RBQL update rule can be generalized to 
\begin{equation}
\label{eq:backwardsQprob}
    Q(S_t,A_t) \leftarrow \sum_{s \in \mathcal{S}_{t+1}}\bigg[(R_{s} + \gamma\ \underset{a}{\text{max}}\ Q(s,a)) \cdot p\bigg]
\end{equation}
where $\mathcal{S}_{t+1}$ is the set of possible states when taking $A_t$ from $S_t$, $p$ is the probability of reaching $s$ when taking $A_t$ from $S_t$ and $R_s$ is the reward of reaching $s$. In a deterministic environment, $\mathcal{S}_{t+1}$ only consists of one state with $p = 1$, negating these additions. Whether RBQL would be as effective in non-deterministic environments as in deterministic environments is something to be explored in further studies.

The only constraint on the algorithm that cannot easily be circumvented is its episodic nature. Because the agent relies on a terminal state from which to propagate the rewards backwards from, a continuous task implementation seems impossible to implement.

\section{Conclusion}
This paper has introduced recursive backwards Q-learning, a model-based reinforcement learning algorithm that evaluates all known state-action pairs of the model at the end of each episode with the Q-learning update rule. It has also shown how recursive backwards Q-learning relates to, adapts and improves on them. This paper has presented an implementation of recursive backwards Q-learning in the Godot game engine to test its performance. Through multiple tests, it has been shown to be superior in finding the shortest path through a randomly generated grid world maze. It has been argued that this algorithm could be adapted to solve other deterministic, episodic tasks more quickly than Q-learning. Further, it has given avenues for further research in adapting recursive backwards Q-learning for non-deterministic problems.

\printbibliography

\end{document}

%% file: diagramme/QProblem/q0.tex
\begin{tikzpicture}[scale=0.65]
    \fill[green] (4,0) rectangle ++ (1,1);  
    \draw[step=1cm,color=black] (0,0) grid (5,1);
    
    \node[scale=0.5] at (.5,.8) {-1};
    \node[scale=0.5] at (.5,.2) {-1};
    \node[scale=0.5] at (.8,.5) {-1};
    \node[scale=0.5] at (.2,.5) {-1};
    
    \node[scale=0.5] at (1.5,.8) {-1};
    \node[scale=0.5] at (1.5,.2) {-1};
    \node[scale=0.5] at (1.8,.5) {-1};
    \node[scale=0.5] at (1.2,.5) {-1};
    
    \node[scale=0.5] at (2.5,.8) {-1};
    \node[scale=0.5] at (2.5,.2) {-1};
    \node[scale=0.5] at (2.8,.5) {-1};
    \node[scale=0.5] at (2.2,.5) {-1};
    
    \node[scale=0.5] at (3.5,.8) {-1};
    \node[scale=0.5] at (3.5,.2) {-1};
    \node[scale=0.5] at (3.8,.5) {-1};
    \node[scale=0.5] at (3.2,.5) {-1};
    
    \node[scale=0.5] at (4.5,.8) {0};
    \node[scale=0.5] at (4.5,.2) {0};
    \node[scale=0.5] at (4.8,.5) {0};
    \node[scale=0.5] at (4.2,.5) {0};
\end{tikzpicture}

%% file: diagramme/QProblem/greedy0.tex
\begin{tikzpicture}[scale=0.65]
    \fill[green] (4,0) rectangle ++ (1,1);  
    \draw[step=1cm,color=black] (0,0) grid (5,1);
    
    \node at (.5,0.5) {\input{diagramme/arrows/omni}};
    \node at (1.5,0.5) {\input{diagramme/arrows/omni}};
    \node at (2.5,0.5) {\input{diagramme/arrows/omni}};
    \node at (3.5,0.5) {\input{diagramme/arrows/omni}};
    
\end{tikzpicture}

%% file: diagramme/QProblem/q1.tex
\begin{tikzpicture}[scale=0.65]
    \fill[green] (4,0) rectangle ++ (1,1);  
    \draw[step=1cm,color=black] (0,0) grid (5,1);
    
    \node[scale=0.5] at (.5,.8) {-1};
    \node[scale=0.5] at (.5,.2) {-1};
    \node[scale=0.5] at (.7,.5) {-1.5};
    \node[scale=0.5] at (.2,.5) {-1};
    
    \node[scale=0.5] at (1.5,.8) {-1};
    \node[scale=0.5] at (1.5,.2) {-1};
    \node[scale=0.5] at (1.7,.5) {-1.5};
    \node[scale=0.5] at (1.2,.5) {-1};
    
    \node[scale=0.5] at (2.5,.8) {-1};
    \node[scale=0.5] at (2.5,.2) {-1};
    \node[scale=0.5] at (2.7,.5) {-1.5};
    \node[scale=0.5] at (2.2,.5) {-1};
    
    \node[scale=0.5] at (3.5,.8) {-1};
    \node[scale=0.5] at (3.5,.2) {-1};
    \node[scale=0.5] at (3.8,.5) {4.5};
    \node[scale=0.5] at (3.2,.5) {-1};
    
    \node[scale=0.5] at (4.5,.8) {0};
    \node[scale=0.5] at (4.5,.2) {0};
    \node[scale=0.5] at (4.8,.5) {0};
    \node[scale=0.5] at (4.2,.5) {0};
\end{tikzpicture}

%% file: diagramme/QProblem/greedy1.tex
\begin{tikzpicture}[scale=0.65]
    \fill[green] (4,0) rectangle ++ (1,1);  
    \draw[step=1cm,color=black] (0,0) grid (5,1);
    
    \node at (.5,0.5) {\input{diagramme/arrows/leftupdown}};
    \node at (1.5,0.5) {\input{diagramme/arrows/leftupdown}};
    \node at (2.5,0.5) {\input{diagramme/arrows/leftupdown}};
    \node at (3.5,0.5) {\input{diagramme/arrows/right}};
    
\end{tikzpicture}

%% file: diagramme/QProblem/q2.tex
\begin{tikzpicture}[scale=0.65]
    \fill[green] (4,0) rectangle ++ (1,1);  
    \draw[step=1cm,color=black] (0,0) grid (5,1);
    
    \node[scale=0.5] at (.5,.8) {-1};
    \node[scale=0.5] at (.5,.2) {-1};
    \node[scale=0.5] at (.7,.5) {-1.3};
    \node[scale=0.5] at (.2,.5) {-1};
    
    \node[scale=0.5] at (1.5,.8) {-1};
    \node[scale=0.5] at (1.5,.2) {-1};
    \node[scale=0.5] at (1.7,.5) {-1.3};
    \node[scale=0.5] at (1.2,.5) {-1};
    
    \node[scale=0.5] at (2.5,.8) {-1};
    \node[scale=0.5] at (2.5,.2) {-1};
    \node[scale=0.5] at (2.7,.5) {0.78};
    \node[scale=0.5] at (2.2,.5) {-1};
    
    \node[scale=0.5] at (3.5,.8) {-1};
    \node[scale=0.5] at (3.5,.2) {-1};
    \node[scale=0.5] at (3.7,.5) {7.25};
    \node[scale=0.5] at (3.2,.5) {-1};
    
    \node[scale=0.5] at (4.5,.8) {0};
    \node[scale=0.5] at (4.5,.2) {0};
    \node[scale=0.5] at (4.8,.5) {0};
    \node[scale=0.5] at (4.2,.5) {0};
\end{tikzpicture}

%% file: diagramme/QProblem/greedy2.tex
\begin{tikzpicture}[scale=0.65]
    \fill[green] (4,0) rectangle ++ (1,1);  
    \draw[step=1cm,color=black] (0,0) grid (5,1);
    
    \node at (.5,0.5) {\input{diagramme/arrows/leftupdown}};
    \node at (1.5,0.5) {\input{diagramme/arrows/leftupdown}};
    \node at (2.5,0.5) {\input{diagramme/arrows/right}};
    \node at (3.5,0.5) {\input{diagramme/arrows/right}};
    
\end{tikzpicture}

%% file: diagramme/QProblem/q3.tex
\begin{tikzpicture}[scale=0.65]
    \fill[green] (4,0) rectangle ++ (1,1);  
    \draw[step=1cm,color=black] (0,0) grid (5,1);
    
    \node[scale=0.5] at (.5,.8) {-1};
    \node[scale=0.5] at (.5,.2) {-1};
    \node[scale=0.5] at (.7,.5) {-1.6};
    \node[scale=0.5] at (.2,.5) {-1};
    
    \node[scale=0.5] at (1.5,.8) {-1};
    \node[scale=0.5] at (1.5,.2) {-1};
    \node[scale=0.5] at (1.7,.5) {-0.8};
    \node[scale=0.5] at (1.2,.5) {-1};
    
    \node[scale=0.5] at (2.5,.8) {-1};
    \node[scale=0.5] at (2.5,.2) {-1};
    \node[scale=0.5] at (2.7,.5) {3.15};
    \node[scale=0.5] at (2.2,.5) {-1};
    
    \node[scale=0.5] at (3.5,.8) {-1};
    \node[scale=0.5] at (3.5,.2) {-1};
    \node[scale=0.5] at (3.7,.5) {8.63};
    \node[scale=0.5] at (3.2,.5) {-1};
    
    \node[scale=0.5] at (4.5,.8) {0};
    \node[scale=0.5] at (4.5,.2) {0};
    \node[scale=0.5] at (4.8,.5) {0};
    \node[scale=0.5] at (4.2,.5) {0};
\end{tikzpicture}

%% file: diagramme/QProblem/greedy3.tex
\begin{tikzpicture}[scale=0.65]
    \fill[green] (4,0) rectangle ++ (1,1);  
    \draw[step=1cm,color=black] (0,0) grid (5,1);
    
    \node at (.5,0.5) {\input{diagramme/arrows/leftupdown}};
    \node at (1.5,0.5) {\input{diagramme/arrows/right}};
    \node at (2.5,0.5) {\input{diagramme/arrows/right}};
    \node at (3.5,0.5) {\input{diagramme/arrows/right}};
    
\end{tikzpicture}

%% file: diagramme/QProblem/q4.tex
\begin{tikzpicture}[scale=0.65]
    \fill[green] (4,0) rectangle ++ (1,1);  
    \draw[step=1cm,color=black] (0,0) grid (5,1);
    
    \node[scale=0.5] at (.5,.8) {-1};
    \node[scale=0.5] at (.5,.2) {-1};
    \node[scale=0.5] at (.7,.5) {-1.66};
    \node[scale=0.5] at (.2,.5) {-1};
    
    \node[scale=0.5] at (1.5,.8) {-1};
    \node[scale=0.5] at (1.5,.2) {-1};
    \node[scale=0.5] at (1.7,.5) {0.52};
    \node[scale=0.5] at (1.2,.5) {-1};
    
    \node[scale=0.5] at (2.5,.8) {-1};
    \node[scale=0.5] at (2.5,.2) {-1};
    \node[scale=0.5] at (2.7,.5) {4.96};
    \node[scale=0.5] at (2.2,.5) {-1};
    
    \node[scale=0.5] at (3.5,.8) {-1};
    \node[scale=0.5] at (3.5,.2) {-1};   
    \node[scale=0.5] at (3.7,.5) {9.32};
    \node[scale=0.5] at (3.2,.5) {-1};
    
    \node[scale=0.5] at (4.5,.8) {0};
    \node[scale=0.5] at (4.5,.2) {0};
    \node[scale=0.5] at (4.8,.5) {0};
    \node[scale=0.5] at (4.2,.5) {0};
\end{tikzpicture}

%% file: diagramme/QProblem/greedy4.tex
\begin{tikzpicture}[scale=0.65]
    \fill[green] (4,0) rectangle ++ (1,1);  
    \draw[step=1cm,color=black] (0,0) grid (5,1);
    
    \node at (.5,0.5) {\input{diagramme/arrows/leftupdown}};
    \node at (1.5,0.5) {\input{diagramme/arrows/right}};
    \node at (2.5,0.5) {\input{diagramme/arrows/right}};
    \node at (3.5,0.5) {\input{diagramme/arrows/right}};
    
\end{tikzpicture}

%% file: diagramme/QProblem/q5.tex
\begin{tikzpicture}[scale=0.65]
    \fill[green] (4,0) rectangle ++ (1,1);  
    \draw[step=1cm,color=black] (0,0) grid (5,1);
    
    \node[scale=0.5] at (.5,.8) {-1};
    \node[scale=0.5] at (.5,.2) {-1};
    \node[scale=0.5] at (.7,.5) {-1.1};
    \node[scale=0.5] at (.2,.5) {-1};
    
    \node[scale=0.5] at (1.5,.8) {-1};
    \node[scale=0.5] at (1.5,.2) {-1};
    \node[scale=0.5] at (1.7,.5) {1.99};
    \node[scale=0.5] at (1.2,.5) {-1};
    
    \node[scale=0.5] at (2.5,.8) {-1};
    \node[scale=0.5] at (2.5,.2) {-1};
    \node[scale=0.5] at (2.7,.5) {6.17};
    \node[scale=0.5] at (2.2,.5) {-1};
    
    \node[scale=0.5] at (3.5,.8) {-1};
    \node[scale=0.5] at (3.5,.2) {-1};
    \node[scale=0.5] at (3.7,.5) {9.66};
    \node[scale=0.5] at (3.2,.5) {-1};
    
    \node[scale=0.5] at (4.5,.8) {0};
    \node[scale=0.5] at (4.5,.2) {0};
    \node[scale=0.5] at (4.8,.5) {0};
    \node[scale=0.5] at (4.2,.5) {0};
\end{tikzpicture}

%% file: diagramme/QProblem/greedy5.tex
\begin{tikzpicture}[scale=0.65]
    \fill[green] (4,0) rectangle ++ (1,1);  
    \draw[step=1cm,color=black] (0,0) grid (5,1);
    
    \node at (.5,0.5) {\input{diagramme/arrows/leftupdown}};
    \node at (1.5,0.5) {\input{diagramme/arrows/right}};
    \node at (2.5,0.5) {\input{diagramme/arrows/right}};
    \node at (3.5,0.5) {\input{diagramme/arrows/right}};
    
\end{tikzpicture}

%% file: diagramme/QProblem/q6.tex
\begin{tikzpicture}[scale=0.65]
    \fill[green] (4,0) rectangle ++ (1,1);  
    \draw[step=1cm,color=black] (0,0) grid (5,1);
    
    \node[scale=0.5] at (.5,.8) {-1};
    \node[scale=0.5] at (.5,.2) {-1};
    \node[scale=0.5] at (.7,.5) {-0.15};
    \node[scale=0.5] at (.2,.5) {-1};
    
    \node[scale=0.5] at (1.5,.8) {-1};
    \node[scale=0.5] at (1.5,.2) {-1};
    \node[scale=0.5] at (1.7,.5) {3.27};
    \node[scale=0.5] at (1.2,.5) {-1};
    
    \node[scale=0.5] at (2.5,.8) {-1};
    \node[scale=0.5] at (2.5,.2) {-1};
    \node[scale=0.5] at (2.7,.5) {6.93};
    \node[scale=0.5] at (2.2,.5) {-1};
    
    \node[scale=0.5] at (3.5,.8) {-1};
    \node[scale=0.5] at (3.5,.2) {-1};
    \node[scale=0.5] at (3.7,.5) {9.83};
    \node[scale=0.5] at (3.2,.5) {-1};
    
    \node[scale=0.5] at (4.5,.8) {0};
    \node[scale=0.5] at (4.5,.2) {0};
    \node[scale=0.5] at (4.8,.5) {0};
    \node[scale=0.5] at (4.2,.5) {0};
\end{tikzpicture}

%% file: diagramme/QProblem/greedy6.tex
\begin{tikzpicture}[scale=0.65]
    \fill[green] (4,0) rectangle ++ (1,1);  
    \draw[step=1cm,color=black] (0,0) grid (5,1);
    
    \node at (.5,0.5) {\input{diagramme/arrows/right}};
    \node at (1.5,0.5) {\input{diagramme/arrows/right}};
    \node at (2.5,0.5) {\input{diagramme/arrows/right}};
    \node at (3.5,0.5) {\input{diagramme/arrows/right}};
    
\end{tikzpicture}